\documentclass{article}

\usepackage[nonatbib,preprint]{neurips_2022}

\author{%
Guodong Cao $^{1}$\quad
 Zhibo Wang $^{1, 2}$ \quad
 Xiaowei Dong $^{1}$ \quad
 Zhifei Zhang  $^{3}$
 \vspace{-1cm}
 \And Hengchang Guo $^{1}$  \qquad \quad
 Zhan Qin $^{2}$ \qquad \quad
Kui Ren $^{2}$\\ \\
 $^{1}$ School of Cyber Science and Engineering, Wuhan University, P. R. China.  \\
 $^{2}$ School of Cyber Science and Technology, Zhejiang University, P. R. China \\
  $^{3}$ Adobe Research\\
}

\usepackage[utf8]{inputenc} 
\usepackage[T1]{fontenc}    
\usepackage{hyperref}       
\usepackage{url}            
\usepackage{booktabs}       
\usepackage{amsfonts}       
\usepackage{nicefrac}       
\usepackage{microtype}      
\usepackage{xcolor}         

\usepackage{graphicx}
\usepackage{algorithm}
\usepackage{autobreak}
\usepackage{algpseudocode}
\usepackage{amsmath}
\usepackage{amssymb}
\usepackage{float}
\usepackage{wrapfig}
\usepackage{subfigure}
\usepackage{caption}
\usepackage{colortbl}
\usepackage{diagbox}
\usepackage[capitalize]{cleveref}
\definecolor{mygray}{gray}{.9}

\usepackage{colortbl}

\newcommand{\proposed}{{VFD-Adv}}

\title{Vanilla Feature Distillation for Improving the Accuracy-Robustness Trade-Off in\\Adversarial Training}

\begin{document}

\maketitle

\begin{abstract}

  Adversarial training has been widely explored for mitigating attacks against deep models. However,  most existing works are still trapped in the dilemma between higher accuracy and stronger robustness since they tend to fit a model towards robust features (not easily tampered with by adversaries) while ignoring those non-robust but highly predictive features. To achieve a better robustness-accuracy trade-off, we propose the Vanilla Feature Distillation Adversarial Training (VFD-Adv), which conducts knowledge distillation from a pre-trained model (optimized towards high accuracy) to guide adversarial training towards higher accuracy, i.e., preserving those non-robust but predictive features. More specifically, both adversarial examples and their clean counterparts are forced to be aligned in the feature space by distilling predictive representations from the pre-trained/clean model, while previous works barely utilize predictive features from clean models. Therefore, the adversarial training model is updated towards maximally preserving the accuracy as gaining robustness. A key advantage of our method is that it can be universally adapted to and boost existing works. Exhaustive experiments on various datasets, classification models, and adversarial training algorithms demonstrate the effectiveness of our proposed method.

\end{abstract}

\section{Introduction}

Deep neural networks (DNNs) have widely dominated various tasks, e.g., image classification\cite{he2016identity,krizhevsky2012imagenet}, object detection\cite{moeslund2001survey,ren2015faster}, autonomous driving\cite{levinson2011towards}, etc. However, DNNs are known to be vulnerable to adversarial examples generated by overlaying carefully designed perturbations onto original/clean examples\cite{goodfellow2014explaining,kurakin2018adversarial,madry2017towards}. A. Ilyas et al.\cite{ilyas2019adversarial} gave an insight to adversarial examples, i.e., rather than “bugs”, one explanation is that, adversarial examples arise as a result of non-robust features tempered by small perturbations which indicate a wrong class.

Against those adversarial attacks, adversarial training is explored to improve the robustness of DNNs, typically by feeding adversarial examples to the model during the training stage\cite{bai2021recent,chen2021cartl,ganin2016domain}. Generally speaking, adversarial training is formulated as a min-max optimization problem, where perturbations are generated to maximize the original loss, and then the model is optimized against the perturbations/attacks by minimizing the loss. A model will become immune to adversarial attacks by forcing the model to focus on robust features, which would robustly correlate true labels to adversarial examples. Although these methods have successfully improved the robustness of deep models, they are mostly trapped in the dilemma between gaining robustness and preserving high accuracy. Recent studies\cite{chen2020robust,cui2021learnable,rade2021helper} found that knowledge distillation can further improve adversarial training. By resorting to third-party models, these methods would learn a more reasonable decision boundary, achieving higher accuracy. However, due to greedily fitting a model towards adversarial examples, they ignored predictive features from clean examples thus limiting model accuracy. Therefore, it is imperative
to explore the predictive features to mitigate accuracy degradation in adversarial training.

In this paper, we propose the Vanilla Feature Distillation Adversarial Training ({\proposed}) to achieve a better trade-off between accuracy and robustness by distilling predictive representations from a highly accurate vanilla model. The basic idea is that preserving those non-robust but predictive features in adversarial training could mitigate accuracy degradation. To this end, we propose to constrain the latent representations from the adversarial training model to be close to that of the vanilla model. Please note that the vanilla model is always fed by clean examples, while the adversarial training model accepts both clean and adversarial examples. In this way, those non-robust but predictive features will be preserved, achieving high accuracy and robustness at the same time in adversarial training. Exhaustive experiments on various datasets, classification models, and adversarial training algorithms demonstrate the effectiveness of our proposed method. In summary, our main contributions are three-fold:

\begin{itemize}
   \item To the best of our knowledge, we give the first attempt to introduce vanilla feature distillation into adversarial training, which achieves a better trade-off between accuracy and robustness.
   \item The proposed vanilla feature distillation adversarial training could be a universally adaptable plug-in for existing related methods to boost accuracy and robustness.
   \item Extensive experiments on diverse classification models, datasets, and adversarial training methods demonstrate the superior performance of our method in terms of accuracy and robustness.
\end{itemize}

\section{Related works}

\subsection{Adversarial Attack}

Since Adversarial attacks were first introduced in \cite{goodfellow2014explaining}, many attack algorithms have been proposed to analyze the vulnerability of deep neural networks. This line of research could be roughly divided into two categories according to the knowledge of the adversary, i.e., white-box attacks and black-box attacks. Some studies focus on white-box attacks, where the adversary has full access to the parameters of the target model. With the full knowledge of the target model, attacks can be conducted by Fast Gradient Sign Method (FGSM)\cite{goodfellow2014explaining}, Momentum Iterative Method (MIM)\cite{dong2018boosting}, Projected Gradient Descent (PGD)\cite{madry2017towards}, and many other methods \cite{tramer2017ensemble,chakraborty2018adversarial}. Besides, several works focus on black-box attacks, in which the adversary cannot access the target model\cite{liu2016delving}. In this case, the adversary carries out the attack with the transferability of adversarial examples, i.e., adversarial examples crafted from one model can effectively attack the other model. Without knowing the parameters of the target model, Carlini Wagner Attack (C$\&$W)\cite{carlini2017towards}, GAP \cite{poursaeed2018generative}, Feature Importance-aware Attack (FIA)\cite{wang2021feature}, etc. can be implemented\cite{bhagoji2018practical,papernot2017practical}. In this paper, we aim to help the target model achieve a better trade-off between accuracy and robustness in both white-box attacks and black-box attacks.

\subsection{Adversarial Training}

In order to defend against adversarial attacks, adversarial training\cite{madry2017towards,wang2019improving, zhang2019theoretically} was proposed to improve the robustness of a model by augmenting the training dataset with adversarial examples when training. Zhang et al.\cite{zhang2019theoretically} theoretically proved a trade-off between robustness and accuracy in adversarial training and proposed TRADES to trade adversarial robustness against accuracy. Based on the TRADES, Wang et al.\cite{wang2019improving} investigated the distinctive influence of misclassified and correctly classified examples and proposed Misclassification Aware adveRsarial Training (MART). These methods have successfully enhanced the robustness of deep models, while both of them failed to provide a model with high robustness and high accuracy. Recently, Some researchers found that adversarial training can be improved with knowledge distillation\cite{ chen2020robust,cui2021learnable,rade2021helper}. Helper-based adversarial training (HAT)\cite{rade2021helper} and Learnable Boundary Guided Adversarial Training (LBGAT)\cite{cui2021learnable} achieved a better trade-off with the knowledge transferred from third-party models. \cite{chen2020robust} used smoothed labels from Knowledge Distillation (KD) to calibrate the notorious overconfidence of logits generated by pre-trained models. However, they only distill from the logits of the teacher model, while it is difficult for the training model to learn those high accuracy features from the teacher model. By contrast, our method inherits the high accuracy knowledge from the vanilla model by approximating the extracted features to the vanilla features, achieving a better trade-off between accuracy and robustness.

\section{Preliminary}

To better understand our proposed method, we will briefly introduce deep neural network, adversarial attack, adversarial training, and knowledge distillation in this section.

\textbf{Deep Neural Networks.} In this paper, we focus on deep neural network based images classification models. A Deep Neural Network $F$ with parameters $\theta$ can be denoted as $F(x;\theta):\mathcal{X} \rightarrow \mathcal{Y}$, where $\mathcal{X}$ is input space and $\mathcal{Y}$ is label space. At layer $l \in \{1,2,...,L\}$, $L$ is the number of layers, we denote the output at layer $l$ as $F^l(x;\theta)$. Usually, the training process of a neural network is to minimize loss function $\mathcal{L} =\frac{1}{N}\sum_{i=1}^{N}\mathcal{J}(F(x_i;\theta), y_i)$ on training data $D$ where $N$ is the number of training instances and $y_i$ is the ground truth of $x_i$, $\mathcal{J}(\cdot,\cdot)$ is usually the cross-entropy for a classification model.

\textbf{Adversarial Attack.} Recent studies show that deep learning models are vulnerable to adversarial examples \cite{yuan2019adversarial}. In this paper, we focus on untargeted adversarial attack. Given a classification model $F(x;\theta)$, the goal of untargeted adversarial attack is to find a small perturbation to generate an adversarial example $x^{adv}$, to mislead the classifier $F(x^{adv};\theta) \neq y$. Typically, the  $\ell_p$-norm of the perturbation should be less than $\epsilon$, i.e. $\left \|x^{adv} - x \right \|_p \leq \epsilon$.

\textbf{Adversarial Training.} Adversarial training is a method for defending against adversarial attacks by augmenting the training dataset with adversarial examples. Adversarial training can be formulated as a min-max optimization problem as follows:
\begin{equation}
\centering
    \arg \; \underset{\theta}{\min} \; \underset{x^{adv}}{\max} \;
    \mathbb{E}_{(x, y)\sim\mathcal{D} } \left[ \mathcal{L}(F\left(x^{adv};\theta\right),y) \right],\;
    s.t.\; \left\|x^{adv}-x\right\|_{p} \le  \epsilon,
    \label{equ:adv_train}
\end{equation}

where $x^{adv}$ is the adversarial example that can maximize loss within $\ell_p$-normXQ distance $\epsilon$, $\theta$ is parameters of model $F$ that needs to be updated to minimize the loss.

\textbf{Knowledge Distillation.} Knowledge distillation is a popular and successful model compression technique that can transfer knowledge from a large pre-trained teacher model to a smaller student model. Given a teacher model $F_t$ with parameters $\theta_t$ and a student model $F_s$ with parameters $\theta_s$, knowledge distillation can be formulated as minimizing a combined loss of soft and hard labels by updating $\theta_s$:

\begin{equation}
\centering
\mathcal{L}_{KD} = \rho \mathcal{L}_{soft}(\boldsymbol{z_s}, \boldsymbol{z_t}) + (1-\rho)\mathcal{L}_{hard}(\boldsymbol{z_s}, y),
\end{equation}

where $\mathcal{L}_{soft}$ and $\mathcal{L}_{hard}$ are metric criterion, $\mathcal{L}_{soft}$ is Kullback-Leibler divergence \cite{hershey2007approximating}, $\mathcal{L}_{hard}$ is cross-entropy, $\boldsymbol{z_s}$ and $\boldsymbol{z_t}$ are logits of $F_s$ and $F_t$ respectively, $y$ is the ground truth label and $\rho$ is a hyper-parameter to control the ratio between $\mathcal{L}_{soft}$ and $\mathcal{L}_{hard}$. After distillation, the student network can achieve better performance than training only with the dataset.

\section{Vanilla Feature Distillation Adversarial Training}
 \label{sec:overview}
In this paper, we propose Vanilla Feature Distillation Adversarial Training ({\proposed}) to realize a better trade-off between robustness and accuracy. This section will first overview the proposed {\proposed} and then describe the detailed design of network and loss functions. Finally, we will further discuss the training strategy of {\proposed}.

\subsection{Overview of {\proposed}}
Since adversarial vulnerability is caused by non-robust and sensitive features (non-semantic and imperceptible features, easily tempered by the adversary), adversarial training serves as a typical method to improve robustness against adversarial attacks by incorporating adversarial examples into the process of model training, forcing the model to focus more on robust features thus become immune to tempered adversarial examples. However, those non-robust features are highly predictive that can be exploited by machines, and omitting them leads to large accuracy degradation on clean examples.

To solve this problem, in this paper, we propose Vanilla Feature Distillation Adversarial Training ({\proposed}) to realize a better trade-off between accuracy and robustness by distilling predictive representations from a high accuracy vanilla model. The basic idea is that, preserving those non-robust but predictive features in adversarial training can mitigate accuracy degradation. To achieve this goal, we constrain features from the adversarial training model that takes adversarial examples and clean examples as input and make them similar to those from the vanilla model fed with the same clean examples. Therefore, those predictive features will be preserved, achieving high accuracy and robustness at the same time.

\begin{figure}[t]
    \centering
    \includegraphics[width=0.75\linewidth]{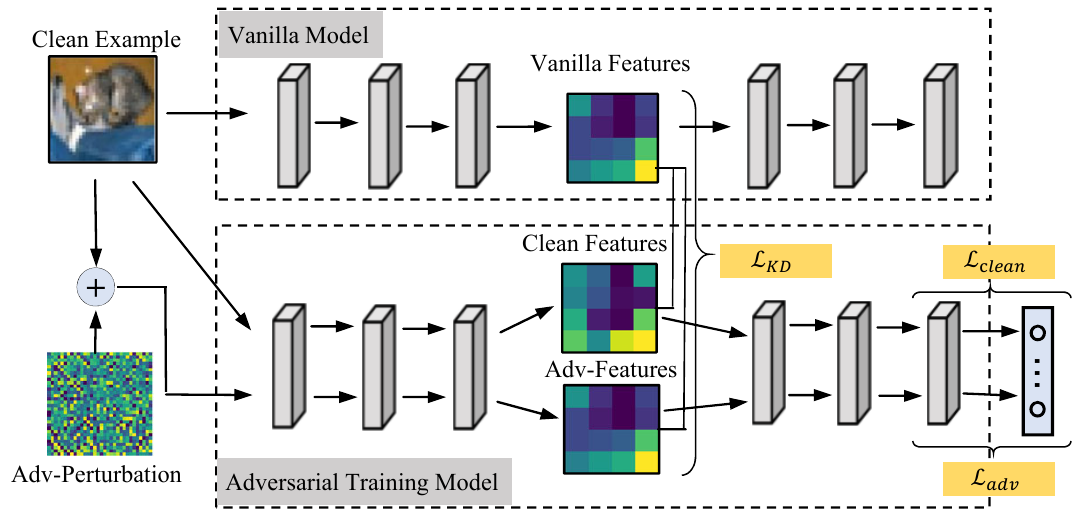}
    \caption{The Overview of the proposed VFD-Adv, which distills predictive representations from the vanilla model to align both adversarial examples and their clean counterparts in the feature space to guide adversarial training towards higher accuracy.}
    \label{fig:overview}
\vspace{-3mm}
\end{figure}

The pipeline of {\proposed} is overviewed in Fig.\ref{fig:overview}, where there are two components: 1) the high accuracy vanilla model $F_{van}$ trained only with merely clean examples, and 2) the adversarial training model $F_{adv}$ that needs to be updated in VFD-Adv. A vanilla model can be one of those off-the-shelf models or pre-trained by ourselves. And the adversarial training model is the model we want after training of {\proposed}. Please note that the vanilla model and the adversarial training model have the same architecture, while parameters of the vanilla model should be freezed once pre-trained. Sharing the spirit of general knowledge distillation and adversarial training during the training process, the vanilla model provides vanilla features for clean examples, while the adversarial training model extracts features for both clean examples and adversarial examples. By matching features to vanilla features, the adversarial training model can learn to preserve those predictive features in adversarial training, thus achieving high accuracy and robustness at the same time.

\subsection{Loss Functions}
In this part, we detail the loss functions of the above mentioned {\proposed}. As illustrated in the overview, we design three loss functions: 1) the clean loss $\mathcal{L}_{clean}$ for supervision of clean examples, and 2) the robustness loss for adversarial training $\mathcal{L}_{adv}$, and 3) the self-distillation loss $\mathcal{L}_{kd}$ for guiding adversarial training towards higher accuracy.

\textbf{The clean loss $\mathcal{L}_{clean}$ : }
In the training of {\proposed}, since our aim is to get a model $F_{adv}$ with high accuracy and high robustness, we take clean examples and their labels for one of supervisions as general adversarial training. This loss can be expressed as:
\begin{equation}
\centering
    \mathcal{L}_{clean} = \frac{1}{N}  \sum_{i=1}^{N}  \mathcal{J}(F_{adv}(x_i;\theta_{adv}),y_i),
\label{L_clean}
\end{equation}
where $x_i$ denotes a clean example, $y_i$ is ground truth for $x_i$, $\mathcal{J}(\cdot,\cdot)$ is cross-entropy for multi-label classification.

\textbf{The robustness loss $\mathcal{L}_{adv}$:}
Meanwhile, we need to improve the robustness of model $F_{adv}$. A typical way to improve robustness is adversarial training, i.e., incorporating adversarial examples into the process of model training. In this way, the adversarial training model learns to correctly classify the adversarial examples. The loss for adversarial training can be expressed as:

\begin{equation}
\begin{aligned}
 \mathcal{L}_{adv} = \frac{1}{N}   \sum_{i=1}^{N} \arg \underset{x_{i}^{adv}}{\max}\mathcal{\phi}  (F_{adv}(x_i^{adv}; \theta_{adv}), \hat{y_i}), \qquad s.t.   \left\|x^{adv}_{i} - x_{i}\right\|_{p} \le  \epsilon,
\label{L_adv}
\end{aligned}
\end{equation}

where $\hat{y_i}$ can be either soft labels (e.g., logits in TRADES\cite{zhang2019theoretically}) or hard labels (ground truth in ALP\cite{kannan2018adversarial}), $\mathcal{\phi}(\cdot,\cdot)$ is a distance metric (e.g., Kullback-Leibler Divergence in TRADES\cite{zhang2019theoretically}, cross-entropy in \cite{kannan2018adversarial}). $\mathcal{L}_{adv}$ itself tends to force the adversarial model to omit non-robust but predictive features of data, reaching a dilemma between gaining more robustness and preserving higher accuracy.

\textbf{The self-distillation loss $\mathcal{L}_{KD}$ : }
The design of $\mathcal{L}_{KD}$ begins with knowledge distillation. Knowledge distillation is one of the most popular techniques used to transfer knowledge from one model to another. In particular, when the architectures are identical, this is called self-distillation. The self-distilled model achieves higher accuracy on held-out data \cite{mobahi2020self,zhang2019your}. The basic idea is that if we constrain features from the adversarial training model that takes adversarial examples and clean examples as input and make them similar to those from the vanilla model fed with the same clean examples, non-robust but predictive features will be preserved during adversarial training thus mitigate accuracy degradation. Therefore, to get those predictive features, we pre-trained a vanilla model that is optimized towards high accuracy only with clean examples. Then, we distill knowledge from its intermediate layer. More specifically, we call this knowledge vanilla features (i.e.,  output at layer $k$ of the model for clean examples in our setting). The design of $\mathcal{L}_{KD}$ is defined as:

\vspace{-0.4cm}
\begin{equation}
\mathcal{L}_{KD} =\frac{1}{N}  \sum_{i=1}^{N} ( \Gamma(F_{adv}^k(x^{adv}_{i};\theta_{adv}), F_{van}^k(x_{i};\theta_{van}))+ \Gamma (F_{adv}^k(x_{i};\theta_{adv}), F_{van}^k(x_{i};\theta_{van}))),
\label{L_kd}
\end{equation}

where $\Gamma(\cdot,\cdot)$ is a distance metric, in this paper we use Euclidean distance. $k$ denotes layer number, $F_{adv}^k$ is the output of adversarial training model at layer $k$, $F_{van}^k$ is the output of clean examples in vanilla model at layer $k$ (i.e., vanilla features). The first term aims to match features of adversarial examples extracted by the adversarial model to clean counterparts in feature space, while the second term will force features of clean examples extracted by the adversarial model to approximate vanilla features. Different from traditional self-distillation, $\mathcal{L}_{KD}$ forces both features of clean examples $F_{adv}^k(x)$ and features of adversarial examples $F_{adv}^k(x^{adv})$ to match vanilla features, thus adversarial model will not omit non-robust features and mitigate accuracy degradation.

\subsection{Training Strategy}
In this part, we demonstrate the training strategy of {\proposed}. Instead of proposing a new adversarial training paradigm, our method is a plug-in for existing adversarial training methods. Based on ~\cref{L_clean},~\cref{L_adv} and~\cref{L_kd}, we first pre-train a high accuracy vanilla model $F_{van}$ with only clean examples, which architecture is the same as the adversarial training model $F_{adv}$. Then, we get the output at the $k_{th}$ layer of clean examples as vanilla features $F^{k}_{van}(x)$. At the same time, the corresponding output for adversarial training model of adversarial examples $F^{k}_{adv}(x_{adv})$ and clean examples $F^{k}_{adv}(x)$ are extracted. Then, we update the model under the supervision of $\mathcal{L}_{clean}$ to achieve better accuracy, adversarial training loss $\mathcal{L}_{adv}$ to improve robustness and self-distillation loss $\mathcal{L}_{kd}$ to mitigate accuracy degradation. Overall, the objectives of {\proposed} can be formulated as follows:

\vspace{-0.42cm}
\begin{equation}
    \mathcal{L}_{total} = \mathcal{L}_{clean}+\beta \cdot \mathcal{L}_{adv}+\lambda \cdot \mathcal{L}_{KD},
\end{equation}

where hyperparameters $\beta$ and $\lambda$ are used to adjust the contribution of each term in the total loss. A large value of $\beta$ will force the model to update towards minimizing loss on adversarial examples, and the robustness of the model will increase while the accuracy on clean examples will decrease. $\lambda$ controls feature difference between extracted features and vanilla features. Detailed training algorithm of {\proposed} can be found in Algorithm \ref{Algorithm 1}.

\renewcommand{\algorithmicrequire}{\textbf{Input:}}
\renewcommand{\algorithmicensure}{\textbf{Output:}}
\begin{algorithm}
\caption{Training of {\proposed}}     
\label{Algorithm 1}       
\begin{algorithmic}[1] 
\Require  Clean examples $x$,  adversarial examples  $x^{adv}$,  model parameters $\theta$ , lables $y$ , adversarial training model $F_{adv}$ ,  vanilla model $F_{van}$,   the model layer output $F^k(\cdot)$, perturbation radius $\varepsilon$,   update amplitude $\alpha$, data set $mathcal{ D}$, number of epochs $N$,   hyperparameter $\lambda$ and $\beta$, learning rate $\eta$. 
\Ensure  Model  $F_{adv}(\theta_{adv})$   
  \\ Initialize the adversarial training model $F_{adv}$ and freeze vanilla model $F_{van}$
  \\For {epoch = 1 to E:}
   \\ \quad For {batch $(x, y)$ in $D$:}
   \\ \qquad Extract features of clean examples $F_{van}^k(x;\theta_{van})$ from $F_{van}$ given $x$
    \\ \qquad   $ x^{adv} = { max}  \mathcal{J}(F_{adv}(x;\theta_{adv}),y)$
    \\ \qquad   Clip $x^{adv}$ to meet $\left \| x^{adv}-x \right \|_2\le  \varepsilon $
    \\ \qquad $ \mathcal{L}_{clean} = \mathcal{J}(F_{adv}(x;\theta_{adv}),{y})$
    \\ \qquad $ \mathcal{L}_{adv} = \mathcal{\phi}(F_{adv}(x^{adv};\theta_{adv}),{y})$
    \\ \qquad Extract features of clean examples $F_{adv}^k(x;\theta_{adv})$ from model $F_{adv}$ given $x$
    \\ \qquad Extract features of adversarial examples $F_{adv}^k(x^{adv};\theta_{adv})$ from model $F_{adv}$ given $x^{adv}$
    \\ \qquad  $\mathcal{L}_{KD} = \left \|F_{van}^k(x;\theta_{van})- F_{adv}^k(x;\theta_{adv}) \right \| _{2}+\left \| F_{van}^k(x;\theta_{van})- F_{adv}^k(x^{adv};\theta_{adv})\right \| _{2} $
    \\ \qquad  $\mathcal{L}_{total} =   \mathcal{L}_{clean}+\beta \cdot \mathcal{L}_{adv}+\lambda \cdot \mathcal{L}_{KD}$
    \\ \qquad  $ \theta_{adv} = \theta_{adv}-\eta \nabla_{\theta_{adv}}  \mathcal{L}_{total} $
    \\ \quad EndFor
 \\EndFor
\end{algorithmic}
\end{algorithm}

\section{Experiments}

\subsection{Experimental setting}

\textbf{Datasets and Classifiers.} For a fair comparison, we follow the previous works\cite{wang2019improving,cui2021learnable,chen2020robust} to conduct experiments on CIFAR10 and CIFAR100\cite{krizhevsky2009learning} . The CIFAR10 dataset has 10 classes and per class contains 6,000 images, while the CIFAR100 dataset has 100 classes and per class contains 600 images. For each dataset, We split it into one training set and one test set in a ratio of 5:1. In addition, we evaluate our method on ResNet18\cite{he2016identity} and VGG16\cite{simonyan2014very} which are widely used in adversarial robustness benchmarks.

\textbf{Attack Methods.} We consider three typical attack methods, i.e., PGD (\emph{step}=20, $\ell_{\infty}$, $\varepsilon$=8/255), FGSM ($\ell_{\infty}$, $\varepsilon$=8/255) and C$\&$W (\emph{kappa}=0, \emph{step}=1000, \emph{lr}=0.01). Moreover, we evaluate the robustness of the model in two attack settings (i.e., white-box setting and black-box setting), and we use WidResNet\cite{zagoruyko2016wide} as the surrogate model to generate adversarial examples in black-box setting.

\begin{figure}[!b]
\centering
\includegraphics[width=0.8\linewidth]{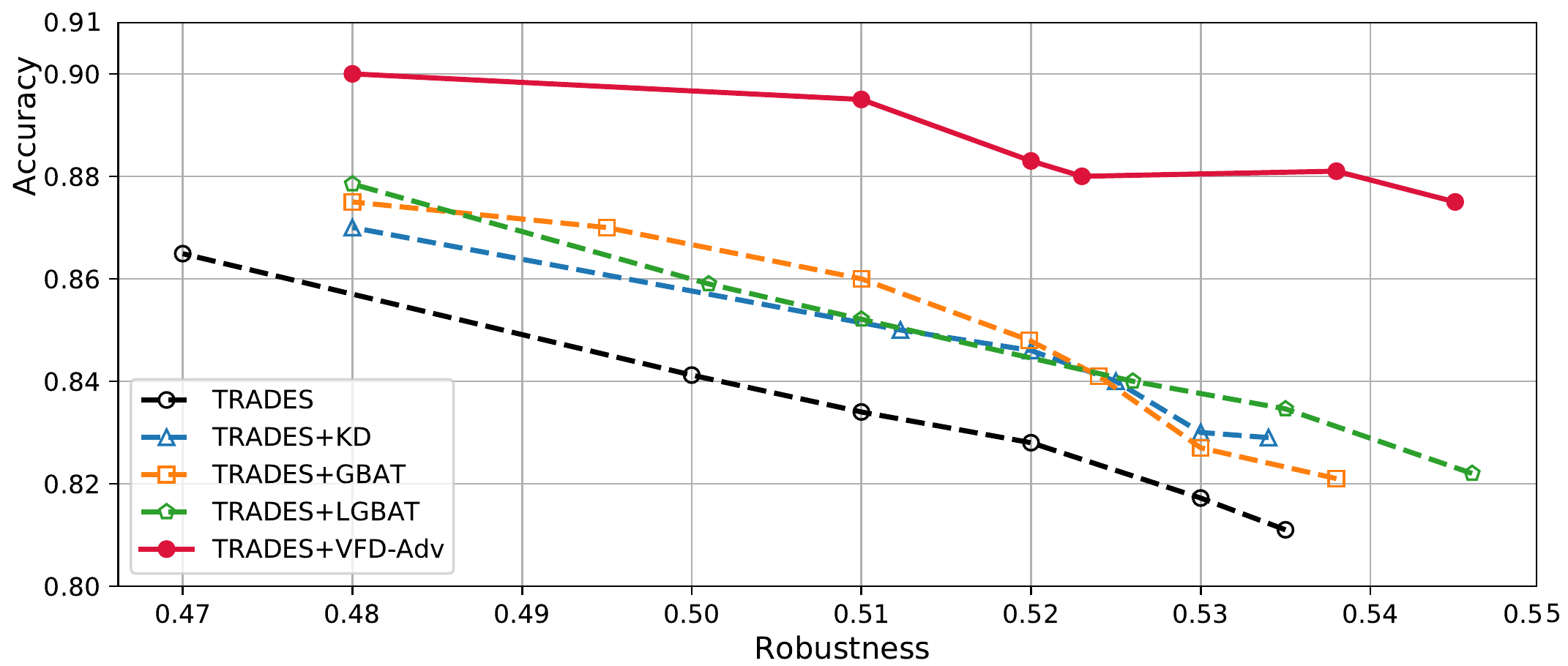}
\caption{Accuracy vs. robustness trade-off of different adversarial training methods (i.e., the original TRADES and several corresponding enhanced versions). From left to right, we increase the Trade-off parameter $\beta$ of TRADES from 1 to 6 with the step of 1.}
\label{fig:roc}
\vspace{-5mm}
\end{figure}

\textbf{Baselines and Parameter Settings.} To verify the effectiveness of the proposed method, we compare it with four related state-of-the-art adversarial training methods (i.e., ALP\cite{kannan2018adversarial}, TRADES\cite{rade2021helper}, MART\cite{wang2019improving}, HAT\cite{rade2021helper}) and three enhanced methods (i.e., KD\cite{chen2020robust}, BGAT and LBGAT\cite{cui2021learnable}) which can further improve the performance. For these methods, we follow their default parameter settings and use the SGD optimizer\cite{cherry1998sgd} with a learning rate of 0.1 and a batch size of 128 to train corresponding robust models.  We conduct all the experiments on a server with one Nvidia Tesla V100 16GB GPU card.
\label{sec:details}


\begin{table}
\begin{center}
\small
  {
   \caption{Robust and clean accuracy of ResNet18 under several white-box and black-box (WidResNet is the surrogate model) attacks on CIFAR10. The best results are highlighted in bold.}
   \label{tab:res_cifar10}
   \setlength{\tabcolsep}{2mm}
    \begin{tabular}{c|c|c|c|c|c|c|c}
    \toprule  
     {\diagbox{DEFENSE}{ATTACK}} &\multicolumn{3}{|c|}{black-box attack}&\multicolumn{3}{|c|}{ white-box attack}&~\\
    ~ & PGD & FGSM & C$\&$W & PGD & FGSM & C$\&$W &  ACC \\
    \toprule  
    {Vanilla} & 8.2\% &   41.1\% &  90.5\%&  0.0\% &  22.0\%&  0.0\%&  94.4\% \\
      \midrule  
     {ALP} & 83.1\% &  82.5\% &  81.4\% &  51.8\% &  55.9\% &  29.7\% & 83.7\%\\
     {ALP+KD}&  82.9\% &  83.0\% &  84.8\% &  51.9\% &  58.4\% &  \textbf{33.1\%} &  84.0\% \\
     {ALP+BGAT}&  83.5\% &  82.5\% &  84.3\% &  52.2\% &  57.3\% &  30.8\% &  84.1\% \\
     {ALP+LBGAT}&  83.3\% &  82.7\% &  83.4\% &  52.1\% &  59.6\% &  31.4\% &  84.5\% \\
     \rowcolor{mygray}
     {ALP+{\proposed}} & \textbf{ 86.1\%}&\textbf{ 86.0\%}&\textbf{ 87.8\% }&\textbf{ 52.6\%}&\textbf{ 61.3\%}&\textbf{ 33.1\%}&\textbf{ 88.3\%}\\

      \midrule  
     {TRADES} & 79.9\% &  79.5\% &  80.1\% &  53.5\% &  56.7\% &  53.2\% &  81.1\% \\
     {TRADES+KD}&  81.1\% &  79.6\% &  80.3\% &  52.8\% &  56.8\% &  50.1\% &  82.1\% \\
     {TRADES+BGAT}&  80.4\% & 81.6\% &  81.3\% &  53.7\% &  57.2\% &   51.3\% &  82.9\% \\
     {TRADES+LBGAT}&  81.2\% &  81.7\% &  82.1\% &  54.1\% &  57.6\% &  52.7\% &  83.1\% \\
     \rowcolor{mygray}
     {TRADES+{\proposed}} & \textbf{ 86.3\%}&\textbf{ 85.4\%}&\textbf{ 86.8\%}&\textbf{ 54.7\%}&\textbf{ 59.8\%}&\textbf{ 55.5\%}&\textbf{ 87.5\%}\\

      \midrule  
      {MART} & 81.0\% &  80.8\% &  82.3\% &  \textbf{55.4\%} &  59.5\% &  46.3\% &  82.3\% \\
     {MART+KD}&  79.9\% &  79.3\% &  81.0\% &  54.9\% &  58.4\% &  45.7\% &  81.1\% \\
     {MART+BGAT}&  81.0\% &  80.5\% &  82.1\% &  54.4\% &  58.7\% &  43.4\% &  82.1\% \\
     {MART+LBGAT}&  80.9\% &  80.4\% &  82.0\% &  55.2\% &  59.2\% &  45.7\% &  82.0\% \\
     \rowcolor{mygray}
     {MART+{\proposed}} & \textbf{ 88.4\%}&\textbf{ 86.8\%}&\textbf{ 87.9\%}& 54.6\%&\textbf{ 61.6\%}&\textbf{ 47.1\%}&\textbf{ 88.7\%}\\
      \midrule  
     {HAT} & 82.5\% &  81.9\% &  83.8\% & 49.8\% &  55.4\% &  44.9\% & 83.9\% \\
     {HAT+KD}&  84.5\% &  84.1\% &  85.8\% &  50.1\% &  56.8\% & 43.1\% &  84.9\% \\
     {HAT+BGAT}&  83.2\% &  82.7\% &  84.2\% &  50.4\% &  56.9\% & 43.8\% &  84.2\% \\
     {HAT+LBGAT}&  84.0\% &  83.2\% &  84.8\% &  50.0\% &  56.1\% &  45.2\% &  84.9\% \\
     \rowcolor{mygray}
     {HAT+{\proposed}} & \textbf{ 86.5\%}&\textbf{ 85.8\%}&\textbf{ 87.6\%}&\textbf{ 52.2\%}&\textbf{ 59.5\%}&\textbf{ 46.5\%}&\textbf{ 88.3\%}\\
       \bottomrule 
    \end{tabular}
    }
\vspace{-6mm}
\end{center}
\end{table}

\subsection{Evaluation on Robustness and Accuracy}

Fig.~\ref{fig:roc} shows the comparison of the accuracy vs. robustness trade-off between the original TRADES and its corresponding enhanced versions. We vary the robustness parameter $\beta$ in TRADES to obtain models with different robustness. We can observe that the models trained by different methods exhibit similar robustness when using the same $\beta$, but our method (i.e., TRADES+{\proposed}) outperforms the others significantly in clean accuracy, achieving a better trade-off between accuracy and robustness.

To quantitatively compare the performance between the proposed {\proposed} and the baselines, we present the robust and clean accuracy of corresponding trained robust ResNet18 under several typical attack methods. Table~\ref{tab:res_cifar10} shows the results on CIFAR10 and Table~\ref{tab:res_cifar100} shows the results on CIFAR100.

As shown in Table~\ref{tab:res_cifar10}, our method significantly improves the robust and clean accuracy of the original adversarial training methods in most cases (white-box attack or black-box attack). For example, compared with the original TRADES, the TRADES+{\proposed} (i.e., equipped with our proposed method) can further improve the robust accuracy by 4.3\% and clean accuracy by 6.4\% on average.  Meanwhile, the TRADES+{\proposed} also outperforms those state-of-the-art methods (e.g., equipped with LBGAT, etc.) 3.1\% in robust accuracy and 4.4\% in clean accuracy on average, achieving the best performance. Moreover, we can draw the same conclusions when combining our method with other adversarial training methods (i.e., ALP, MART, HAT), which indicates the scalability and effectiveness of our method.

\begin{table}[t]
\small
\begin{center}
  {
   \caption{Robust and clean accuracy of ResNet18 under several white-box and black-box (WidResNet is the surrogate model) attacks on CIFAR100. The best results are highlighted in bold.  }
   \label{tab:res_cifar100}
   \setlength{\tabcolsep}{2mm}
    \begin{tabular}{c|c|c|c|c|c|c|c}
    \toprule  
     {\diagbox{DEFENSE}{ATTACK}} & \multicolumn{3}{|c|}{ black-box attack}&\multicolumn{3}{|c|}{ white-box attack}&~\\
     ~ & PGD & FGSM & C\&W & PGD  & FGSM & C\&W &  {ACC}\\
    \toprule  
    {Vanilla} &  17.4\% &  18.9\% & 69.9\% &0.0\% & 7.3\% & 0.0\%& 75.2\% \\
      \midrule  
     {ALP} &50.5\% & 51.6\% & 51.2\% & 30.0\% & 31.7\% & \textbf{23.2\%} & 51.4\% \\
     {ALP+KD}& 55.0\% & 54.9\% & 54.5\% & 30.4\% & 29.8\% & 20.1\% & 56.0\% \\
     {ALP+BGAT}& 52.1\% & 49.5\% & 50.2\% & 30.4\% & \textbf{33.9\%} & 21.1\% & 56.3\% \\
     {ALP+LBGAT}& 59.3\% & 58.5\% &58.1\% & 30.5\% & 31.1\% & 22.0\% & 59.4\% \\
     \rowcolor{mygray}
     {ALP+{\proposed}} &\textbf{62.3\%}&\textbf{61.6\%}&\textbf{62.9\%}&\textbf{31.8\%}&33.0\%&\textbf{23.2\%}&\textbf{63.2\%}\\

      \midrule  
     {TRADES} &55.0\% & 54.4\% & 55.9\% & 25.4\% & 28.0\% & 24.5\% & 56.0\% \\
     {TRADES+KD}& 56.2\% & 55.5\% & 57.4\% & 29.3\% & 31.7\% & 27.3\% & 57.4\% \\
     {TRADES+BGAT}& 59.6\% & 59.0\% & 61.1\% & 29.6\% & 32.8\% & 27.5\% & 61.2\% \\
     {TRADES+LBGAT}& 54.7\% & 55.9\% & 57.6\% & 28.9\% & 32.5\% & 26.3\% & 57.6\% \\
     \rowcolor{mygray}
     {TRADES+{\proposed}} &\textbf{64.8\% }& \textbf{63.1\%} & \textbf{67.2\%} & \textbf{30.1\% }& \textbf{34.1\%} & \textbf{27.7\%} & \textbf{67.3\% }\\

      \midrule  
     {MART} &53.9\% & 53.1\% & 54.8\% & 32.2\% & 34.0\% & 28.2\% & 54.9\% \\
     {MART+KD}& 46.7\% & 46.4\% & 47.6\% & 30.3\% & 31.3\% & \textbf{30.4\%} & 56.6\% \\
     {MART+BGAT}&57.3\% & 56.3\% & 58.5\% & 29.8\% & 32.7\% & 29.9\% & 58.6\% \\
     {MART+LBGAT}& 56.1\% & 55.5\% & 57.2\% & 32.4\% & 33.5\% & 30.3\% & 57.2\% \\
     \rowcolor{mygray}
     {MART+{\proposed}} &\textbf{62.1\%} & \textbf{62.9\% }& \textbf{62.4\%} & \textbf{32.5\%} & \textbf{35.4\%} & \textbf{30.4\%} & \textbf{63.8\% }\\
      \midrule  
     {HAT} &57.5\% & 56.8\% & 59.0\% & 26.5\% & 29.5\% & \textbf{20.2\%} & 59.0\% \\
     {HAT+KD}& 58.2\% & 57.6\% & 59.2\% & 25.3\% & 29.0\% & 19.0\% & 59.2\% \\
     {HAT+BGAT}& 60.4\% & 59.5\% & 61.8\% & 27.2\% & 30.8\% & 19.1\% & 61.8\% \\
     {HAT+LBGAT}& 60.1\% & 59.4\% & 60.1\% & 27.7\% & 30.0\% & 18.7\% & 61.2\% \\
     \rowcolor{mygray}
     {HAT+{\proposed}} &\textbf{62.7\%} & \textbf{62.4\% }& \textbf{63.1\%} & \textbf{28.0\%} & \textbf{32.8\%} & 20.1\% & \textbf{64.5\% }\\
     \bottomrule 
    \end{tabular}
    }
\vspace{-0.6cm}
\end{center}
\end{table}

In addition, we also present the comparison results on a more complex classification task (i.e., CIFAR100) in Table~\ref{tab:res_cifar100}. The results show that our method still can improve the robust accuracy by 2.3\% and clean accuracy by 4.5\% on average compared with the other state-of-the-art methods, demonstrating the superior performance and the generality of our proposed method on different datasets. In conclusion, the proposed method mitigates the accuracy degradation significantly and improve the robustness. Meanwhile, the proposed method also can be adapted to existing related works flexibly as an extra effective regularization item. We show the experimental results of VGG16 in supplementary file, which show better improvement compared to that on ResNet18.

\begin{figure}[!b]
\vspace{-0.5cm}
\begin{center}
    \includegraphics[width=0.9\linewidth]{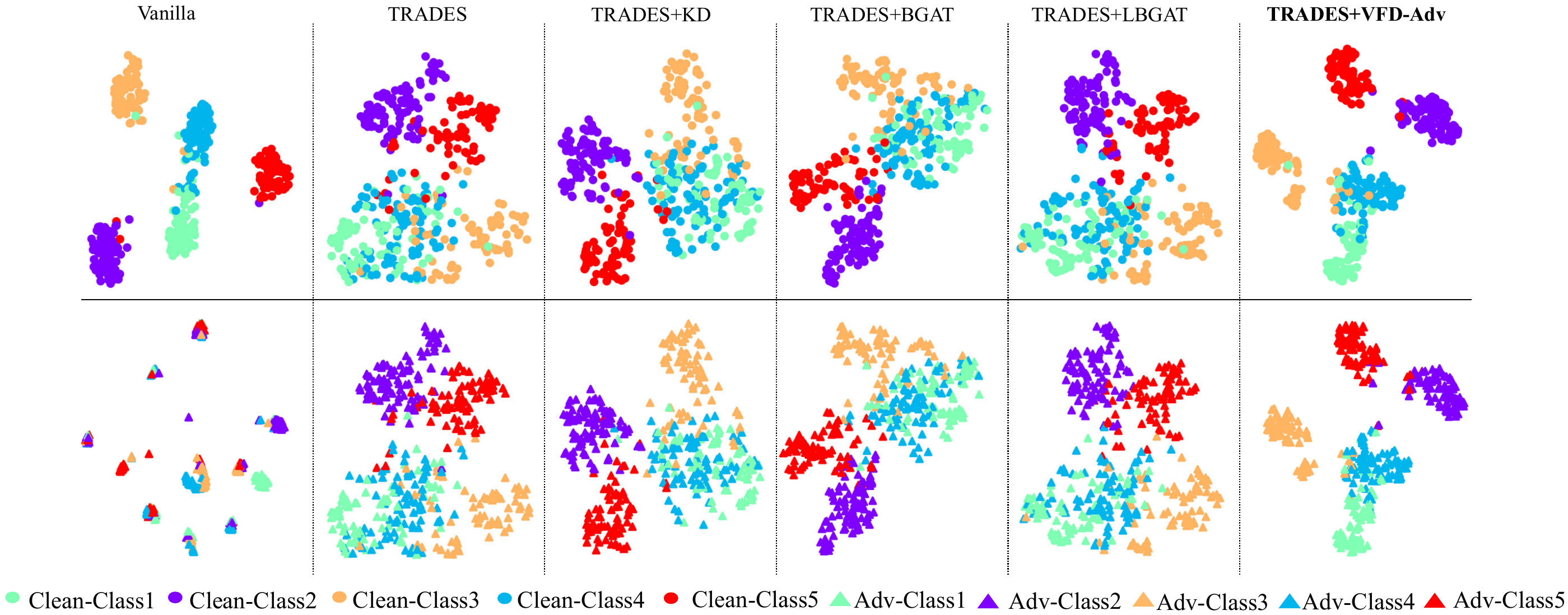}
    \caption{Feature visualization of clean examples and adversarial examples in CIFAR10 on ResNet18. Each color represents a class. Dots and triangle represent clean and adversarial examples respectively.}
    \label{fig:T-SNE}
\end{center}
\vspace{-0.5cm}
\end{figure}

\subsection{Feature Visualization Analysis}
Furthermore, to qualitatively analyze the effectiveness of our proposed {\proposed}, we randomly select 5 classes in CIFAR10 and utilize t-SNE\cite{van2008visualizing} to visualize features of 500 clean examples per class and corresponding adversarial examples generated by PGD. As shown in Fig.~\ref{fig:T-SNE}, the first row presents the t-SNE results of clean examples and the second row presents the results of corresponding adversarial examples. For the vanilla model, we can observe that there are 5 clear separable clusters in the case of clean examples but 10 clusters in the case of adversarial examples. The phenomenon indicates that the vanilla model cannot extract right discriminate features from these adversarial examples hence classify them into wrong classes. For all existing defense methods (e.g., TRADE), features of adversarial examples are separated into 5 clusters to some degree, but we can observe that there would appear unexpected obscure boundaries in clean examples, thus leading to degradation in clean accuracy. In contrast, as shown in the most right column, our model can force features of both adversarial examples and clean examples of the adversarial model close to vanilla features, thus realizing clear boundaries in these clusters for both clean examples and adversarial examples. These results are consistent with the rationality of the proposed method.

\subsection{Effect of Parameters in {\proposed}}
\begin{figure}[t]
\centering
\subfigure[]{
\includegraphics[width=0.45\columnwidth]{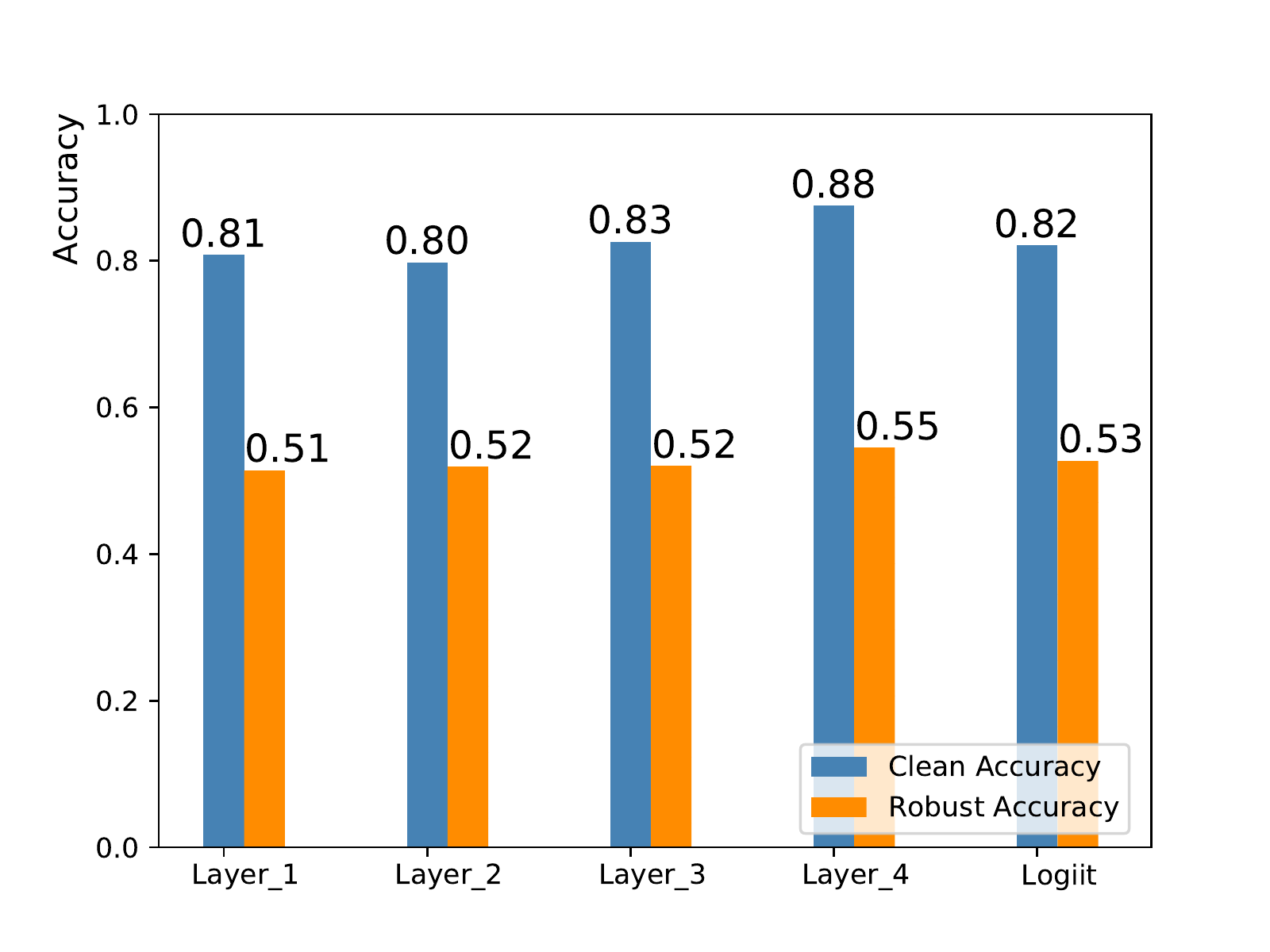}
\label{fig:Layer_acc}}
\hspace{0.01in}
\subfigure[]{
\includegraphics[width=0.46\columnwidth]{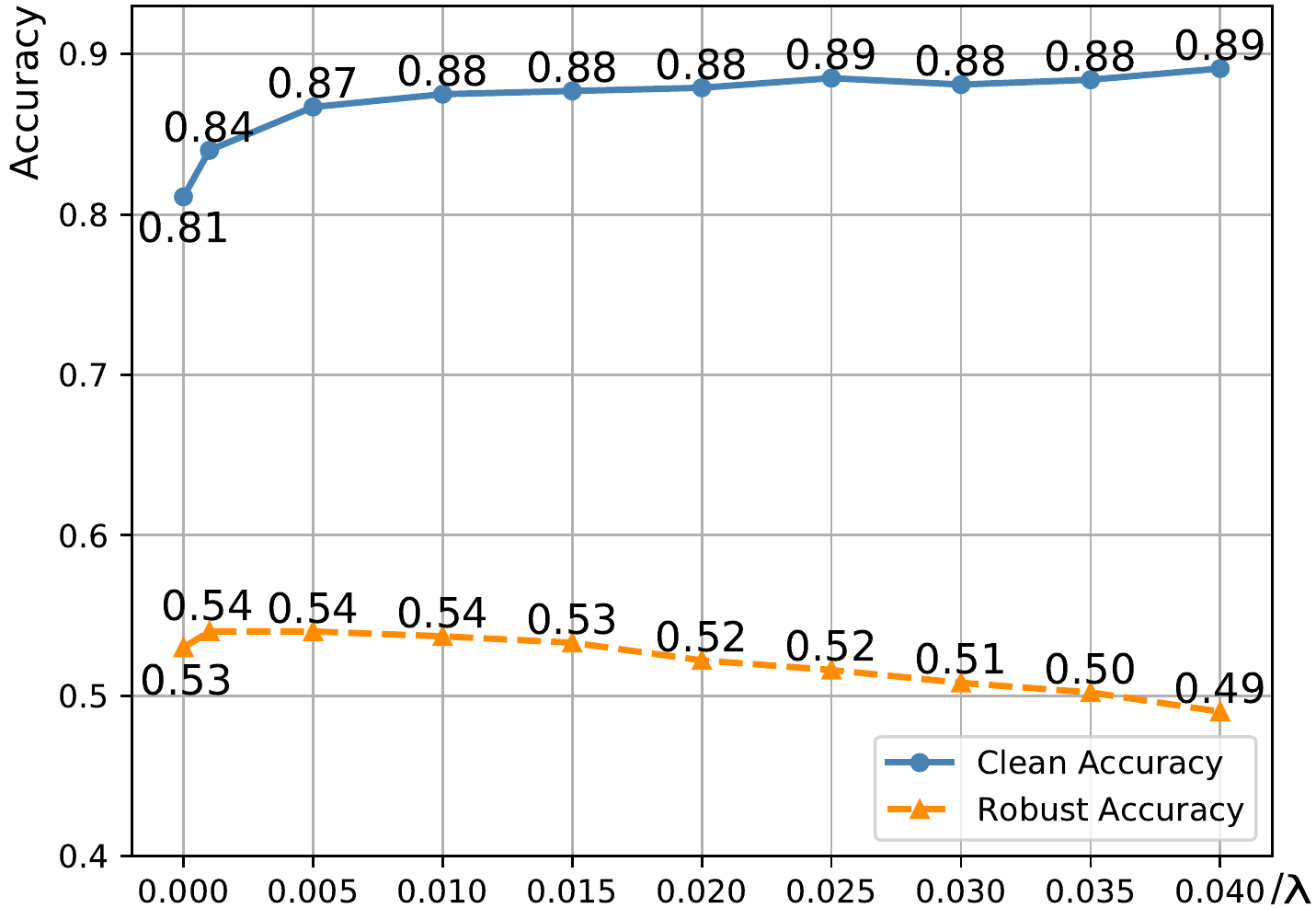}
\label{fig:lamuda_1}
}
\vspace{-0.2cm}
\caption{(a) Robust and clean accuracy of ResNet18 trained by our proposed method (i.e., TRADES+{\proposed}) with distilling the output of different layers in the corresponding vanilla model. (b) Robust and clean accuracy of ResNet18 with different distillation parameter $\lambda$.}
\vspace{-0.5cm}
\end{figure}

There are two parameters, i.e., the choice of vanilla feature layer $k$ and the distillation parameter $\lambda$, which affect the performance of the proposed {\proposed}. Here, we adopt ResNet18 trained on CIFAR10 as the vanilla model to explore the effects of the two parameters. Since ResNet18 has four residual blocks, we choose the output of the last Conv layer of each block (e.g., Layer\_1, etc.) and logit layer respectively, and modify $\lambda$ from 0 to 0.04 with a step of 0.005. As shown in Fig.~\ref{fig:Layer_acc}, we can observe that the proposed method would achieve the best performance in both robust and clean accuracy when using the layer of the last residual block (i.e., Layer\_4).
Compared to early layers, the Layer\_4 contains more high-level and task-specific information while keeping more spatial information than the logit features, hence the Layer\_4 comes as a better choice.
In addition, the distillation parameter $\lambda$ also influences the robust and clean accuracy significantly.
The distillation parameter $\lambda$ is expected to guide the robust model to learn more predictive features like the vanilla model, but also make the robust model less robust due to relying on such non-robust features.
Therefore, as shown in Fig.~\ref{fig:lamuda_1}, a larger $\lambda$ tends to yield higher clean accuracy but diminishes the robustness while a smaller $\lambda$ would achieve better robustness with compromising the clean accuracy. 

\section{Conclusion}
In this work, we proposed Vanilla Feature Distillation Adversarial Training ({\proposed}) to achieve a better trade-off between accuracy and robustness. {\proposed} conducts knowledge distillation from a high-accuracy pre-trained model to guide the adversarial training model to learn more predictive features, which allows the model maximally preserve the accuracy and gain robustness as well. Specifically, the proposed method can be flexibly adapted to and boost existing adversarial training methods. Extensive experiments conducted across different datasets, network architectures, and adversarial training algorithms demonstrate the state-of-the-art performance of our method. We hope that our method can serve as the adversarial robustness benchmark and inspire the community to further ameliorate the accuracy-robustness trade-off.
\label{sec:limitation}

\textbf{Broader Impact and Limitations:} It is crucial to train a robust model against adversarial attacks for deep learning models. While adversarial training generates a robust model, it also causes a decrease in the model's accuracy. This paper proposes Vanilla Feature Distillation Adversarial Training (VFD-Adv) to improve adversarial training and effectively increase the accuracy of robust models, but also introduces a little bit more time for training compared to typical adversarial training. Our work gives a worthwhile effort for further adversarial robustness research and has no known socially detrimental effects. 

\end{document}